\documentclass[letterpaper]{article} % DO NOT CHANGE THIS
\usepackage{aaai25}  % DO NOT CHANGE THIS
\usepackage{times}  % DO NOT CHANGE THIS
\usepackage{helvet}  % DO NOT CHANGE THIS
\usepackage{courier}  % DO NOT CHANGE THIS
\usepackage[hyphens]{url}  % DO NOT CHANGE THIS
\usepackage{graphicx} % DO NOT CHANGE THIS
\usepackage{natbib}  % DO NOT CHANGE THIS AND DO NOT ADD ANY OPTIONS TO IT
\usepackage{caption} % DO NOT CHANGE THIS AND DO NOT ADD ANY OPTIONS TO IT
\usepackage{algorithm}
\usepackage{algorithmic}

\usepackage{newfloat}
\usepackage{listings}
\DeclareCaptionStyle{ruled}{labelfont=normalfont,labelsep=colon,strut=off} % DO NOT CHANGE THIS
\floatstyle{ruled}
\newfloat{listing}{tb}{lst}{}
\floatname{listing}{Listing}
\title{HSEvo: Elevating Automatic Heuristic Design with Diversity-Driven Harmony Search and Genetic Algorithm Using LLMs}
\author {
    Pham Vu Tuan Dat\textsuperscript{\rm 1},
    Long Doan\textsuperscript{\rm 2},
    Huynh Thi Thanh Binh\textsuperscript{\rm 1}
}
\affiliations {
    \textsuperscript{\rm 1}Hanoi University of Science and Technology, Hanoi, Viet Nam\\
    \textsuperscript{\rm 2}George Mason University, Virginia, United States\\
    dat.pvt210158@sis.hust.edu.vn, 
    ldoan5@gmu.edu, 
    binhht@soict.hust.edu.vn
}
\usepackage{bibentry}
\usepackage{amsmath}
\usepackage{multirow}
\usepackage{tcolorbox}
\usepackage{xcolor}
\lstdefinestyle{mypython}{
    language=Python,
    basicstyle=\ttfamily\scriptsize,
    keywordstyle=\bfseries\color{purple},
    commentstyle=\color{green!50!black},
    stringstyle=\color{orange},
    frame=lines,
    showstringspaces=false,
    numbers=left,
    numberstyle=\color{gray},
    breaklines=true,
    tabsize=4,
}
\begin{document}

\maketitle

\begin{abstract}
Automatic Heuristic Design (AHD) is an active research area due to its utility in solving complex search and NP-hard combinatorial optimization problems in the real world. The recent advancements in Large Language Models (LLMs) introduce new possibilities by coupling LLMs with evolutionary computation to automatically generate heuristics, known as LLM-based Evolutionary Program Search (LLM-EPS). While previous LLM-EPS studies obtained great performance on various tasks, there is still a gap in understanding the properties of heuristic search spaces and achieving a balance between exploration and exploitation, which is a critical factor in large heuristic search spaces. In this study, we address this gap by proposing two diversity measurement metrics and perform an analysis on previous LLM-EPS approaches, including FunSearch, EoH, and ReEvo. Results on black-box AHD problems reveal that while EoH demonstrates higher diversity than FunSearch and ReEvo, its objective score is unstable. Conversely, ReEvo's reflection mechanism yields good objective scores but fails to optimize diversity effectively. With this finding in mind, we introduce HSEvo, an adaptive LLM-EPS framework that maintains a balance between diversity and convergence with a harmony search algorithm. Through experimentation, we find that HSEvo achieved high diversity indices and good objective scores while remaining cost-effective. These results underscore the importance of balancing exploration and exploitation and understanding heuristic search spaces in designing frameworks in LLM-EPS.
\end{abstract}

\begin{links}
    \link{Code}{https://github.com/datphamvn/HSEvo}
\end{links}

\begin{figure*}[t]
\centering
\includegraphics[width=0.85\textwidth]{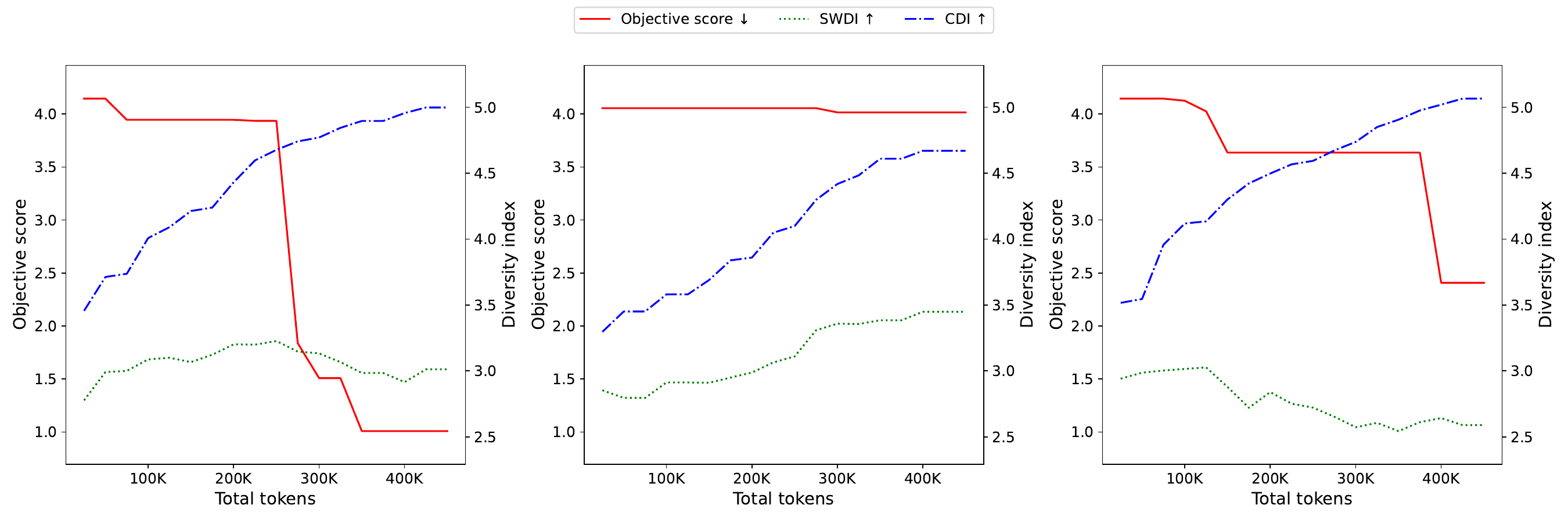} % Reduce the figure size so that it is slightly narrower than the column.
\caption{Diversity indices and objective scores of ReEvo framework on BPO problem through different runs.}
\label{fig:di-and-obj}
\end{figure*}

\section{Introduction}
Heuristics are widely utilized to address complex search and NP-hard combinatorial optimization problems (COPs) in real-world systems. Over the past few decades, significant efforts have been made to develop efficient heuristics, resulting in the creation of many meta-heuristic methods such as simulated annealing, tabu search, and iterated local search.
% \cite{kirkpatrick1983optimization}
% \cite{glover1998tabu}
% \cite{lourencco2003iterated}
% , among others (Martí et al., 2018)
These meticulously crafted methods have been effectively applied across various practical systems.

Nevertheless, varied applications with distinct constraints and goals may necessitate different algorithms or algorithm setups. The manual creation, adjustment, and configuration of a heuristic for a specific problem can be extremely laborious and requires substantial expert knowledge. This is a bottleneck in many application domains. To address this issue, Automatic Heuristic Design (AHD) has been proposed aims to selects, tunes, or constructs effective heuristics for a given problem class automatically \cite{choong2018automatic}. Various approaches have been used in AHD, including Hyper-Heuristics (HHs) \cite{pillay2018hyper} and Neural Combinatorial Optimization (NCO) \cite{qu2020general}. However, HHs are still constrained by heuristic spaces that are predefined by human experts. Additionally, NCO faces limitations related to the necessity for effective inductive bias \cite{drakulic2024bq}, and challenges regarding interpretability and generalizability \cite{liu2023good}.

Recently, the rise of Large Language Models (LLMs) has opened up new possibilities for AHD. It is believed that LLMs \cite{nejjar2023llms,austin2021program} could be a powerful tool for generating new ideas and heuristics. However, standalone LLMs with prompt engineering can be insufficient for producing novel and useful ideas beyond existing knowledge \cite{mahowald2024dissociating}. Some attempts have been made to coupling LLMs with evolutionary computation to automatically generate heuristics, known as LLM-based Evolutionary Program Search (LLM-EPS) \cite{liu2024large,meyerson2024language, chen2024evoprompting}. Initial works such as FunSearch \cite{romera2024funsearch} and subsequent developments like Evolution of Heuristic (EoH) \cite{liu2024eoh} and Reflective Evolution (ReEvo) \cite{ye2024reevo} have demonstrated significant improvements over previous approaches, generating quality heuristics that often surpass current methods. Even so, ReEvo yields state-of-the-art and competitive results compared with evolutionary algorithms, neural-enhanced meta-heuristics, and neural solvers.

A key difference between LLM-EPS and classic AHD lies in the search spaces of heuristics. Classic AHD typically operates within well-defined mathematical spaces such as $R^n$, whereas LLM-EPS involves searching within the space of functions, where each function represents a heuristic as a program. LLM-EPS utilizes LLMs within an evolutionary framework to enhance the quality of generated functions iteratively. Therefore, it is crucial to study and understand the search spaces of heuristics to establish foundational theories and principles for the automatic design of algorithms. This aspect has been largely unconsidered in previous LLM-EPS frameworks.

% In this paper, we explore the characteristics and properties of heuristic search spaces in LLM-EPS through analysis of populations generated at different stages by different frameworks. Besides, we introduce a diversity index inspired by the Shannon entropy index to monitor population diversity and represent the objective scores of the population throughout various evolutionary stages. Finally, building on these foundational principles, a new framework was proposed called HSEvo, which incorporates a new stage called Harmony Search and enhances existing phases (initialization, crossover, mutation) to optimize objective scores, cost efficiency, exploration, and exploitation in heuristic search spaces.

In this paper, we introduce two diversity metrics inspired by the Shannon entropy to monitor population diversity. We also explore relationships between our proposed diversity metrics with objective performance of LLM-EPS frameworks on different optimization problems. Through our analysis, we gain an understanding on the characteristics and properties of heuristic search spaces in LLM-EPS frameworks.  Finally, building on these foundational principles, we propose a new framework called \textbf{HSEvo}, which incorporates a new components based on harmony search algorithm \cite{Shi2012} and enhances other components, including initialization, crossover, and mutation, to optimize objective scores and diversity metrics.

In summary, our contributions are as follows:
\begin{itemize}
\item Two diversity measurement metrics, the Shannon–Wiener Diversity Index and the Cumulative Diversity Index, are used to evaluate the evolutionary progress of populations within the LLM-EPS framework.
% \item Proposed methods to integrate heuristic search into existing LLM-EPS frameworks to enhance diversity metrics and optimize performance. 
\item A novel framework, HSEvo, that aims to balance between the diversity and objective performance to improve the optimization process.
\end{itemize}

\section{Background and Related Works}

\subsection{LLM-based Evolutionary Program Search}
Recent advances in LLM-EPS have shown promising results in AHD. Evolutionary methods have been adopted in both code generation \cite{nejjar2023llms,ma2023eureka,hemberg2024evolving} and text generation \cite{guo2023connecting,yuksekgonul2024textgrad}. Notable among these methods are FunSearch, EoH, and ReEvo. FunSearch employs an island-based evolution strategy, leveraging LLMs like Codey and StarCoder to evolve heuristics for mathematical and COPs, outperforming traditional methods on tasks such as the cap set and admissible set problems. EoH utilizes genetic algorithm with Chain of Thought prompt engineering, consistently achieving superior performance in the traveling salesman and online bin packing problems. ReEvo introduces a reflective component to the evolution process, employing two instances of GPT-3.5 to generate and refine heuristics, which has demonstrated effectiveness across various optimization tasks. These methods highlight the potential of integrating LLMs with evolutionary strategies to enhance the efficiency and effectiveness of AHD solutions.

\subsection{Diversity in Evolutionary Computation}
Diversity plays a pivotal role in enhancing the efficiency of algorithms in multi-objective optimization within the domain of evolutionary computation \cite{SWDI-GA}. Numerous studies have investigated various methods to measure and maintain diversity within populations, as it critically impacts the convergence and overall performance of these algorithms \cite{SWDI-GA2,Wang2012}. Among these, the application of Shannon entropy has been particularly prominent in quantifying diversity and predicting the behavior of genetic algorithms. However, to the best of our knowledge, no existing studies have thoroughly explored diversity within the context of LLM-EPS. This gap in the literature motivates our in-depth exploration of diversity in LLM-EPS frameworks.

\begin{figure*}[t]
\centering
\includegraphics[width=0.85\textwidth]{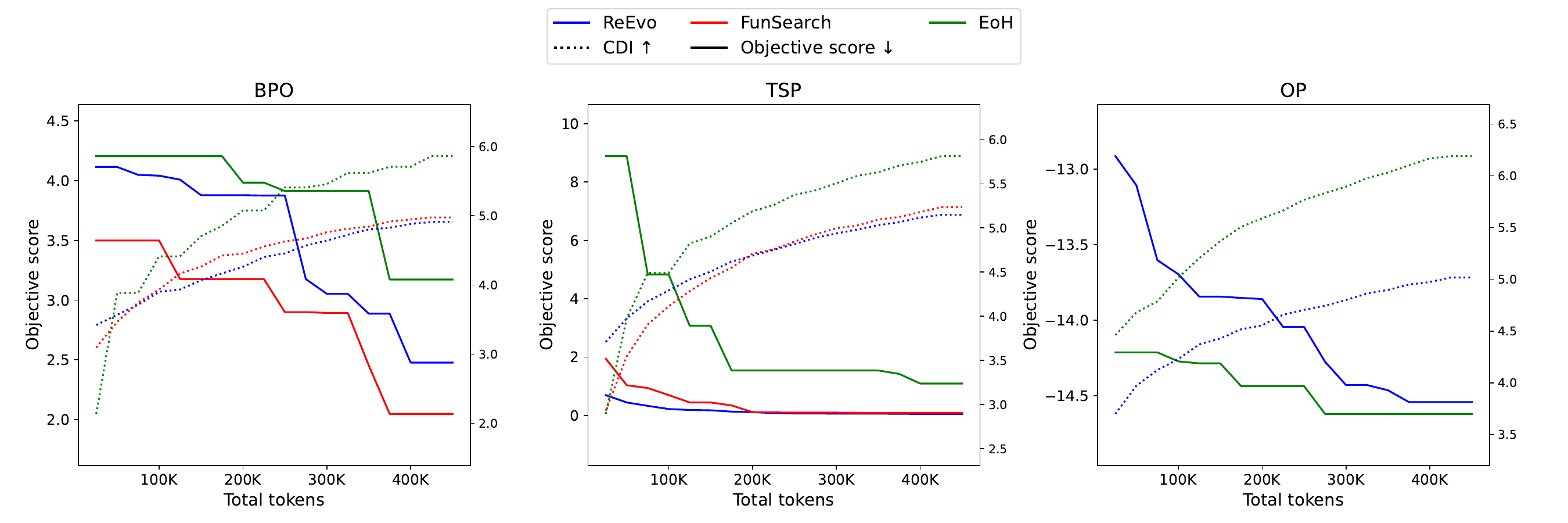} % Reduce the figure size so that it is slightly narrower than the column.
\caption{CDI and objective scores of previous LLM-EPS on different AHD problems.}
\label{fig:analysis-llm-eps}
\end{figure*}

\section{Diversity measurement metrics}
In this section, we introduce a method to encode LLM-EPS population and propose two diversity measurement metrics, Shannon-Wiener diversity index (SWDI) and cummulative diversity index (CDI). We also conduct a diversity analysis on previous LLM-EPS frameworks, FunSearch, EoH and ReEvo.

\subsection{Population encoding}
\label{ssec:pop_encode}
One particular problem of measuring diversity in LLM-EPS is how to encode the population. While each individual in traditional evolutionary algorithm is encoded as a vector, in LLM-EPS they are usually presented as a string of code snippet/program. This poses a challenge in applying previous diversity metrics on population of LLM-EPS frameworks.
To tackle this issue, we suggest an encoding approach consists of three steps: 
(i) removing comments and docstrings using abstract-syntax tree, (ii) standardizing code snippets into a common coding style (e.g., PEP8\footnote{https://peps.python.org/pep-0008}), (iii) convert code snippets to vector representations using a code embedding model.

\subsection{Shannon–Wiener Diversity Index} 
Inspired by ecological studies, SWDI \cite{SWDI} provides a quantitative measure of species diversity within a community. In the context of search algorithms, this index aims to quantify the diversity of the population at any given time step based on clusters of individuals. To compute the SWDI for a given set of individuals within a community, or archive, first, we need to determine the probability distribution of individuals across clusters. This is represented as:
\begin{equation}
    \label{eq:probability_SWDI}
    p_i = \frac{|C_i|}{M}
\end{equation}
where $C_i$ is a cluster of individuals and $M$ represents the total number of individuals across all clusters $\{C_1, \dots, C_N\}$. The SWDI, $H(X)$, is then calculated using the Shannon entropy: 
\begin{equation}
    \label{eq:entropy_SWDI}
    H(X) = - \sum_{i=1}^{N} p_i \log(p_i)
\end{equation}

This index serves as a crucial metric in maintaining the balance between exploration and exploitation within heuristic search spaces. A higher index score suggests a more uniform distribution of individuals across the search space, thus promoting exploration. Conversely, a lower index score indicates concentration around specific regions, which may enhance exploitation but also increase the risk of premature convergence.

To obtain SWDI score, at each time step $t$, we add all individuals $R_i = \{r_1, \dots, r_n\}$ generated at time step $t$
% generated between time intervals $t_i$ and $t_j$ 
to an archive.
% The process can be described by the following equations:
% $$
% P(r_i) = c_i 
% $$
% where $P$ is the pre-processing function, $c_i \in \mathcal{C}$ and $\mathcal{C}$ is the space of all possible code snippets.
% $$
% E(c_i) = v_i 
% $$
% where $E$ is the code embedding model that maps code snippets to a high-dimensional vector space, $v_i \in \mathcal{V}$ is the embedding vector of $c_i$ and $\mathcal{V} = \mathcal{R}^n$ is the $n$-dimensional space of embeddings.
We encode the archive using our proposed population encoding in Section \ref{ssec:pop_encode} to obtain their vector representations $\mathcal{V} = \{v_1, \dots, v_n\}$. After that, we compute the similarity between two embedding vectors $v_i$ and $v_j$ using cosine similarity:
$$
\text{similarity}(v_i, v_j) = \frac{v_i \cdot v_j}{\|v_i\| \|v_j\|}
$$

To find clusters in the archive, we consider each embedding vector $v_i$. We assign $v_i$ to a cluster $C_i$ if the similarity between $v_i$ and all members in $C_i$ is greater than a threshold $\alpha$. Mathematically, $v_i$ is assigned to $C_i$ if
$$
\text{similarity}(v_i, v_k) \geq \alpha \quad \forall k \in \{1, \dots, |C_i|\}
$$
If no cluster $C_i \in \{C_1, \dots, C_N\}$ can satisfy the above condition, we create a new cluster $C_{N+1}$ and assign $v_i$ to $C_{N+1}$. Finally, we compute SWDI score by computing Eq. (\ref{eq:probability_SWDI}), (\ref{eq:entropy_SWDI}) across found clusters.
% where $\alpha$ is the similarity threshold used to determine cluster membership.

\begin{figure*}[t]
\centering
\includegraphics[width=0.8\textwidth]{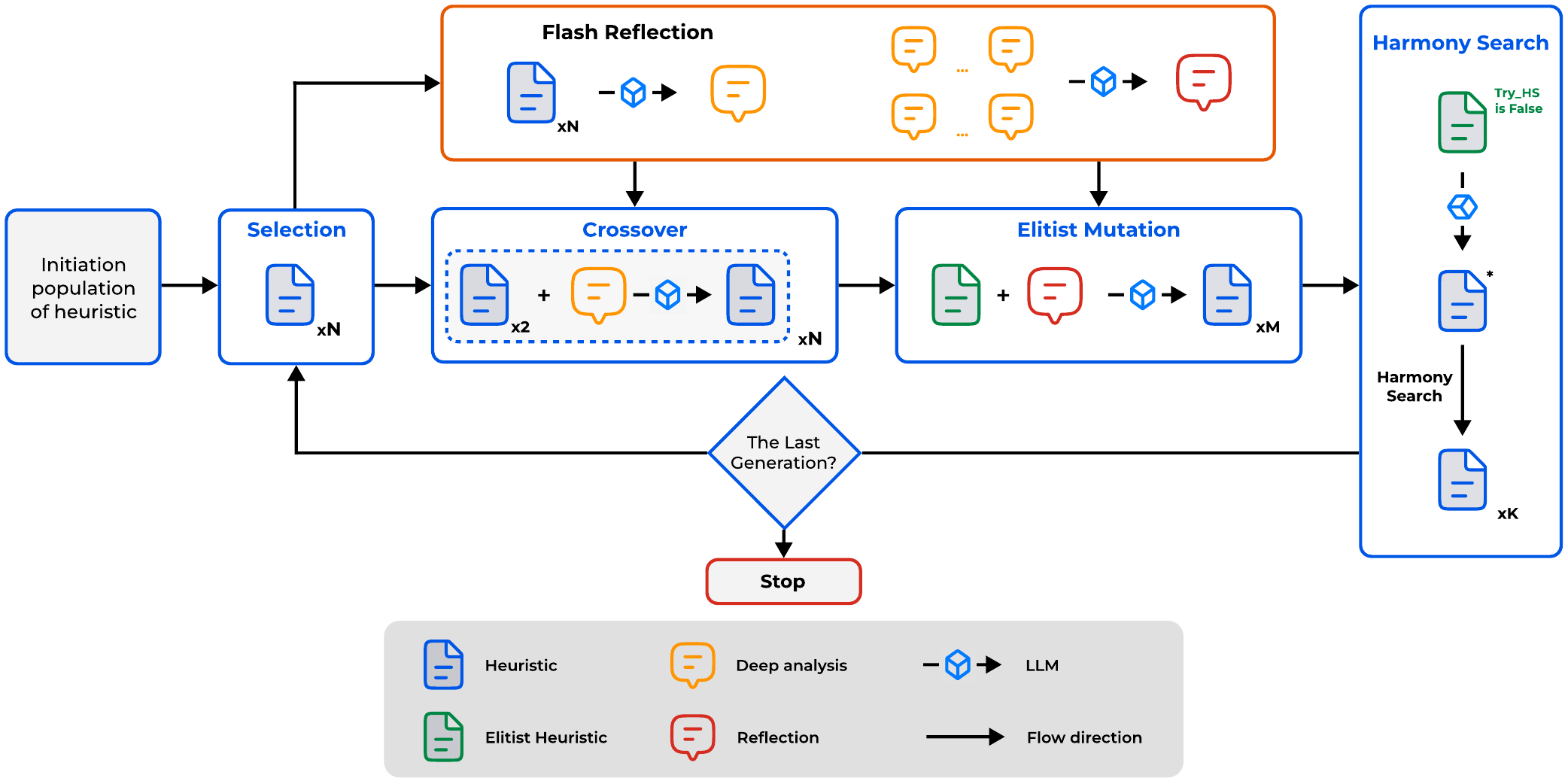} % Reduce the figure size so that it is slightly narrower than the column.
\caption{Overview of the HSEvo framework.}
\label{fig:hsevo_pipline}
\end{figure*}

\begin{figure*}[ht]
    \centering
    \begin{minipage}[t]{0.40\textwidth}
        \centering
        \begin{lstlisting}[style=mypython]
import numpy as np

def heuristics_v2(prize: np.ndarray, 
distance: np.ndarray, maxlen: float) -> np.ndarray:
    reward_distance_ratio = prize / distance
    cost_penalty = np.exp(-distance)

    heuristics = (prize * prize[:, np.newaxis])
    heuristics[distance > maxlen] = 0

    return heuristics
        \end{lstlisting}
        \textbf{(a)} Origin Code
    \end{minipage}
    \hfill
    \begin{minipage}[t]{0.58\textwidth}
        \centering
        \begin{lstlisting}[style=mypython]
import numpy as np

def heuristics_v2(prize: np.ndarray, 
distance: np.ndarray, maxlen: float, 
reward_threshold: float = 0, 
distance_threshold: float = 0, 
cost_penalty_weight: float = 1) -> np.ndarray:
    reward_distance_ratio = prize / distance
    cost_penalty = np.exp(-distance)
    
    heuristics = (prize * prize[:, np.newaxis]) / (distance * distance) * cost_penalty
    heuristics[(distance > maxlen) | (reward_distance_ratio < reward_threshold) | (distance < distance_threshold)] = 0
    
    return heuristics

parameter_ranges = {
    'reward_threshold': (0, 1),
    'distance_threshold': (0, 100),
    'cost_penalty_weight': (0, 2)
}

        \end{lstlisting}
        \textbf{(b)} Modified Code using LLMs
    \end{minipage}
    \caption{An example of how Harmony search component works in HSEvo.}
    \label{fig:app-exp-hs}
\end{figure*}

\subsection{Cumulative Diversity Index}
While SWDI focuses on diversity of different groups of individuals, the Cumulative Diversity Index (CDI) plays a crucial role in understanding the spread and distribution of the whole population within the search space \cite{jost2006entropy}. In the context of heuristic search, the CDI measures how well a system’s energy, or diversity, is distributed from a centralized state to a more dispersed configuration. 
% This metric is particularly useful in assessing the performance of search algorithms where both exploration and exploitation must be balanced to avoid premature convergence while ensuring efficient optimization.

To calculate the CDI, we also consider all individuals within an archive, represented by their respective embeddings. We construct a minimum spanning tree (MST) that connects all individuals within the archive \( A \), where each connection between individuals is calculated using Euclidean distance. This MST provides a structure to assess the diversity of the population. Let \( d_i \) represent the distance of an edge within the MST, where \( i \in \{1, 2, \ldots, \#A - 1\} \). The probability distribution of these distances is given by:

\[
p_i = \frac{d_i}{\sum_{j=1}^{\#A-1} d_j}
\]

The cumulative diversity is then calculated using the Shannon entropy:

\[
H(X) = - \sum_{i=1}^{\#A-1} p_i \log(p_i)
\]

This approach allows us to capture the overall diversity of the search space, providing insights into the spread of solutions. Higher CDI values indicate a more distributed and diverse population, which is essential for maintaining a robust search process. 

An interesting point in Shannon diversity index (SDI) theory is that normalizing the SDI with the natural log of richness is equivalent to the evenness value $\frac{H'}{\ln(S)}$ where \(H'\) represents the richness of SDI and \(S\) is the total number of possible categories \cite{heip1998indices}. The significance of evenness is to demonstrate how the proportions of \(p_i\) are distributed. Now, consider normalizing with the two current diversity indices. First, with SWID, if we have \(N\) clusters and the number of individuals in each cluster from \(C_1\) to \(C_N\) is equal, then evenness will always be \(1\), regardless of the value of \(N\). This is not the desired behavior since we consider the number of clusters to be a component of diversity. Similarly, normalizing CDI will also ignore the impact of the number of candidate heuristics, which correspond to the nodes in the MST. This leads to the proposal not to normalize the SWID and CDI metrics in order to achieve a \([0,1]\) bound.

% In this study, we apply the CDI to analyze the performance of various LLM-EPS frameworks. Our findings suggest that while some approaches, such as EoH, achieve higher diversity indices, they struggle with stability in objective scores. In contrast, approaches like ReEvo demonstrate better objective performance but at the cost of reduced diversity. By integrating the CDI into our analysis, we aim to better understand the trade-offs between diversity and convergence in the design of LLM-EPS frameworks, ultimately leading to more effective and balanced heuristic search strategies.

\subsection{Exploring the correlation between objective score and diversity measurement metrics}
\label{ssec:analysis_cdi_swdi_obj}
To examine the correlation between the two diversity measurement metrics and objective score, we conducted three experimental runs using ReEvo on bin-packing online problem \cite{seiden2002online}. The experiment details can be found in the Appendix. The objective scores and the two diversity metrics from these runs are presented in Fig. \ref{fig:di-and-obj}. Additional results for other tasks are also available in the Appendix.

% The results in Fig.\ref{di-and-obj} yielded two key observations. 
From the figure, there are a few observations that we can drawn.
First, the two diversity measurement metrics have a noticeable correlation on the objective scores of the problem. 
When the objective score is converged into a local optima, the framework either try to focus on the exploration by increasing the diversity of the population, through the increase of SWDI in first and second run, or try to focus on the exploitation of current population, thus decrease the SWDI in the third run. We can also see that focusing on exploration can lead to significant improvement on the objective score, while focusing on exploitation can make the population stuck in local optima for a longer time. However, if the whole population is not diverse enough, as can be seen with the low CDI of the second run, the population might not be able to escape the local optima.

% Second, the trajectory of the Shannon–Wiener Diversity Index emerges as a crucial metric in maintaining the balance between exploration and exploitation during the corresponding experimental runs.

\subsection{Diversity analysis on previous LLM-EPS framework}
\label{ssec:analysis_llm_eps}
% While SWDI plays an important role in understanding the diversity of the population during a single run, it may not be helpful for quantifying the diversity across multiple runs.

To analyse the overall diversity of previous LLM-EPS frameworks, we conduct experiments on three different LLM-EPS frameworks, including FunSearch \cite{romera2024funsearch}, ReEvo \cite{ye2024reevo}, and EoH \cite{liu2024eoh}, on three distinct AHD problems, bin-packing online (BPO), traveling salesmen problem with guided local search solver (TSP) \cite{voudouris1999guided}, and orienteering problem with ACO solver (OP) \cite{ye2024deepaco}. The details of each experiment are presented in the Appendix.  Note that, as highlighted in the previous section, SWDI focuses on understanding the diversity of the population during a single run. As such, it may not be helpful for quantifying the diversity across multiple experiment runs. Fig. \ref{fig:analysis-llm-eps} presents the experiment results.

% Our findings reveal that higher CDI often correlate with better objective scores, with an exception of EoH. 
In BPO and TSP, EoH obtain the highest CDI but got the worst objective score. This implies that EoH 
% heavily suffered from the trade-off between diversity and objective performance.
does not focus enough on exploitation to optimize the population better.
In contrast, while ReEvo and FunSearch obtain lower CDI than EoH on BPO and TSP, they achieve a better objective performance on all three problems. 
The experiments on BPO and TSP problems highlight the inherent trade-off between diversity and objective performance of LLM-EPS frameworks. However, on OP problem, we can see that a high CDI is required to obtain a better objective score, which align with our findings in Section \ref{ssec:analysis_cdi_swdi_obj}.
% while methods such as EoH demonstrate greater diversity, they often suffer from instability in objective scores. In contrast, frameworks like ReEvo, which prioritize convergence, exhibit lower diversity scores, highlighting the inherent trade-off between diversity and objective performance.

\begin{table*}[htb!]
    \centering
    \begin{tabular}{l|c}
        \textbf{Description of Setting} & \textbf{Value} \\
        \hline
        LLM (generator and reflector) & gpt-4o-mini-2024-07-18 \\
        LLM temperature (generator and reflector) & 1 \\
        Maximum budget tokens & 425K tokens \\
        Population size (for EoH, ReEvo and HSEvo) & 30 (initial stage), 10 (other stages) \\
        Mutation rate (for ReEvo and HSEvo) & 0.5 \\
        \# of islands, \# of samples per prompt (for FunSearch) & 10, 4 \\
        Number of independent runs per experiment & 3 \\
        HS size, HMCR, PAR, bandwidth, max iterations (for Harmony Search) & 5, 0.7, 0.5, 0.2, 5 \\
        \hline
        Maximum evaluation time for each heuristic  & 100 sec (TSP-GLS), 50 sec (Other) \\
        % (to cope with invalid heuristics, such as infinite loops) & 50 sec (Other) \\
        \hline
    \end{tabular}
    \caption{Summary of parameters settings.}
    \label{tab:parameters-settings}
\end{table*}

\begin{table*}[]
\centering
\begin{tabular}{|c|c|c|c|c|c|c|}
\hline
\multicolumn{1}{|c|}{\multirow{2}{*}{Method}} & \multicolumn{2}{|c|}{BPO} & \multicolumn{2}{|c|}{TSP} & \multicolumn{2}{|c|}{OP} \\
\cline{2-7}
\multicolumn{1}{|c|}{} & CDI ($\uparrow$) & Obj. ($\downarrow$) & CDI ($\uparrow$) & Obj. ($\downarrow$) & CDI ($\uparrow$) & Obj. ($\downarrow$) \\
\hline
FunSearch & $4.97 \pm 0.24$ & $2.05 \pm 2.01$ & $5.24 \pm 0.14$ & $0.09 \pm 0.06$ & - & - \\
EoH & $\mathbf{5.86 \pm 0.49}$ & $3.17 \pm 2.97$ & $\mathbf{5.81 \pm 0.23}$ & $1.09 \pm 3.11$ & $\mathbf{6.17 \pm 0.42}$ & $-14.62 \pm 0.22$ \\
ReEvo & $4.91 \pm 0.53$ & $2.48 \pm 3.74$ & $5.15 \pm 0.19$ & $0.05 \pm 0.06$ & $5.02 \pm 0.13$ & $-14.54 \pm 0.21$ \\
\hline
HSEvo (ours) & $5.68 \pm 0.35$ & $\mathbf{1.07 \pm 1.11}$ & $5.41 \pm 0.21$ & $\mathbf{0.02 \pm 0.03}$ & $5.67 \pm  0.41$ & $\mathbf{-14.62 \pm 0.12}$ \\
\hline
\end{tabular}
\caption{Experiment results on different AHD problems.}
\label{tab:hsevo_results}
\end{table*}

\begin{table}[]
\centering
\begin{tabular}{|c|c|c|}
\hline
\multicolumn{1}{|c|}{\multirow{2}{*}{Method}} & \multicolumn{2}{|c|}{OP} \\
\cline{2-3}
\multicolumn{1}{|c|}{} & CDI ($\uparrow$) & Obj. ($\downarrow$) \\
\hline
ReEvo & $5.02 \pm 0.13$ & $-14.54 \pm 0.21$ \\
ReEvo + HS & $ 5.11 \pm 0.27 $ & $-14.58 \pm 0.38$ \\
\hline
HSEvo (ours) & $\mathbf{5.67 \pm  0.41}$ & $\mathbf{-14.62 \pm 0.12}$ \\
\hline
\end{tabular}
\caption{Ablation results on harmony search.}
\label{tab:ablation_hs}
\end{table}

\begin{table}[]
\centering
\begin{tabular}{|c|c|c|}
\hline
\multicolumn{1}{|c|}{\multirow{2}{*}{Method}} & \multicolumn{2}{|c|}{OP} \\
\cline{2-3}
\multicolumn{1}{|c|}{} & CDI ($\uparrow$) & Obj. ($\downarrow$) \\
\hline
ReEvo & $4.43 \pm 0.23$ & $-13.84 \pm 1.13$ \\
ReEvo + F.R. & $4.63 \pm 0.37$ & $\mathbf{-14.36 \pm 0.19}$ \\
\hline
HSEvo (ours) & $ \mathbf{4.77 \pm 0.46} $ & $ -14.07 \pm 0.38 $ \\
\hline
\end{tabular}
\caption{Ablation results on flash reflection.}
\label{tab:ablation_flash_reflection}
\end{table}

\section{Automatic Heuristic Design with HSEvo}
In this section, we propose a novel LLM-EPS framework called \textbf{H}armony \textbf{S}earch \textbf{Evo}lution (HSEvo). HSEvo aim to promote the diversity of the population while still achieve better optimization and alleviate the trade-off between diversity and optimization performance with a individual tuning process based on harmony search. HSEvo also aim to reduce the cost incurred from LLM with an efficient flash reflection component. Figure \ref{fig:hsevo_pipline} illustrates the pipeline of our HSEvo framework. Examples of prompts used in each stage can be found in the Appendix.

% The main idea of HSEvo revolves around combining GA with the Harmony Search algorithm. It utilizes LLMs to explore various heuristic design structures broadly. However, most heuristic functions have parameters like thresholds and weights, which are set by LLMs. Despite LLM being a black box, it often yields unconvincing results in complex computational tasks. The research hypothesizes that while the structure of the heuristic function might be "good," the parameter settings (thresholds and weights) by LLMs lead to "poor" evaluation results. Therefore, there is a need to expand the search space in depth. To achieve this, and based on the study of the Hybrid Genetic Algorithm (A Hybrid Genetic Algorithm Based on Harmony Search and its Improving), this research chooses to integrate Harmony Search with GA.

% At each generation, HSEvo maintains a population of N heuristics, denoted as
% $P = \{h_1, \dots , h_N \}$. Each heuristic $h_i$ is evaluated on a set of problem instances and assigned a fitness value $f(h_i)$.

\textbf{Individual encoding.} 
Follow previous works in LLM-EPS (EoH \cite{liu2024eoh}, ReEvo \cite{ye2024reevo}), HSEvo encodes each individual as a string of code snippet generated by LLMs (Fig. 4a). This encoding method allow the flexibility of each individual, i.e., not constrained by any predefined encoding format, and also make it easier to evaluate on the AHD problem.

% HSEvo optimizes toward best-performing heuristics via an evolutionary process, specifically a GP. It diverges from traditional GPs in that (1) individuals are code snippets generated by LLMs, and (2) individuals are not constrained by any predefined encoding format, except for adhering to a specified function signature. HSEvo are summarised as follows.

\textbf{Initialization.} HSEvo initializes a heuristic population by prompting the generator LLM with task specifications that describe the problem and detail the signature of the heuristic function to be searched. Additionally, to leverage existing knowledge, the framework can be initialized with a seed heuristic function and/or external domain knowledge. We promote diversity by implementing in the prompts with various role instructions (e.g., "You are an expert in the domain of optimization heuristics...", "You are Albert Einstein, developer of relativity theory..."). This approach aims to enrich the heuristic generation process by incorporating diverse perspectives and expertise. 
% To avoid the trade-off between diversity 

\textbf{Selection.} HSEvo randomly selects parent pairs from the population, aiming to maintain the balance between 
% diversity and ensure that a broad spectrum of heuristic strategies is explored, 
exploration and exploitation during our optimization process. The random selection may also counteract premature convergence, which is observed through the SWID trajectory outlined in Section \ref{ssec:analysis_cdi_swdi_obj}. 
% Additionally, alternative techniques, such as the island-based evolution employed in FunSearch, can be adopted to further preserve exploration, particularly when a large number of heuristic generations are feasible.

\textbf{Flash reflection.} Reflections can provide LLMs with reinforcement learning reward signals in a verbal format for code generation tasks, as discussed by \cite{shinn2024reflexion}. Later, \cite{ye2024reevo} also proposed integrating Reflections into LLM-EPS in ReEvo as an approach analogous to providing “verbal gradient” information within heuristic spaces. However, we argue that reflecting on each pair of heuristic parents individually is not generalized and wastes resources. To address these issues flash reflection proposed to alternative Reflection method for LLM-EPS. Flash reflection includes two steps as follows:

\begin{itemize}
    \item At time step $t$, we perform grouping on all individuals that are selected from the selection stage and remove all duplication. After that, we use LLM to perform a deep analysis on the ranking on parent pairs. This take inputs as mixed ranking pairs (i.e., one good performance parent and one worse performance), good ranking pairs (i.e., both good performance), and worse ranking pairs (i.e., both worse performance), then return a comprehensive description on how to improve as a text string.
    \item From the current analysis at time step $t$, we use LLM to compare with previous analysis at time step $t - 1$ and output a guide information, which can be used in subsequent stages.
\end{itemize}

\textbf{Crossover.} In this stage, new offspring algorithms are generated by combining elements of the parent algorithms. The goal is to blend successful attributes from two parents via guide result guide information part of flash reflections. Through this step, HSEvo hopes to produce offspring that inherit the strengths of both, which can potentially lead to better-performing heuristics. The prompt includes task specifications, a pair of parent heuristics, guide information part of flash reflection, and generation instructions.

\textbf{Elitist mutation.} HSEvo uses an elitist mutation strategy, where the generator LLM is tasked with mutation an elite individual—representing the best-performing heuristic—by incorporating insights derived from LLM-analyzed flash reflections. Each mutation prompt includes detailed task specifications, the elite heuristic, deep analysis part of flash reflections, and specific generation instructions. This approach leverages accumulated knowledge to enhance the heuristic, ensuring continuous improvement in performance while preserving the quality of the top solutions.

\textbf{Harmony Search.}  
From the analysis on Section \ref{ssec:analysis_cdi_swdi_obj} and \ref{ssec:analysis_llm_eps}, we hypothesize that if the population is too diverse, each individuals inside will more likely to be not optimized, which may cause harm to the optimization process. We employ the Harmony Search algorithm to alleviate this issue by optimizing the parameters (e.g., thresholds, weights, etc.) of best individuals in the population. The process is as follows:
\begin{itemize}
    \item First, we use LLM to extract parameters from the best individual (i.e., code snippets, programs) of the population and define a range for each parameters (Fig. 4).
    % \item In alignment with the work of \cite{Shi2012}, which demonstrated that hybridizing Genetic Algorithms (GA) with Harmony Search (HS) can enhance solution quality, we are inspired to integrate a Harmony Search framework into the LLM-based Evolutionary Program Search (LLM-EPS) to optimize the relevant parameters in the current problem domain.
    \item Following the work \cite{Shi2012}, we optimize the above parameters with harmony search algorithm.
    \item After parameter optimization, we mark this individual and add it back to the population. All marked individuals will not be optimized again in the future time steps.
\end{itemize} 
% is employed to fine-tune the parameters (thresholds, weights) of the best individual. This phase expands the search space and optimizes the heuristics by systematically adjusting the parameters within defined ranges. 
% By fine-tuning these parameters, we aim to avoid the trade-off between diversity and objective performance existed in EoH. 
Fine-tuning these parameters makes the individual more optimized, therefore allowing us to encourage diversity during the prompting while avoid the trade-off between objective performance and diversity, as can be seen with EoH.

\begin{figure*}[ht]
\centering
\includegraphics[width=0.85\textwidth]{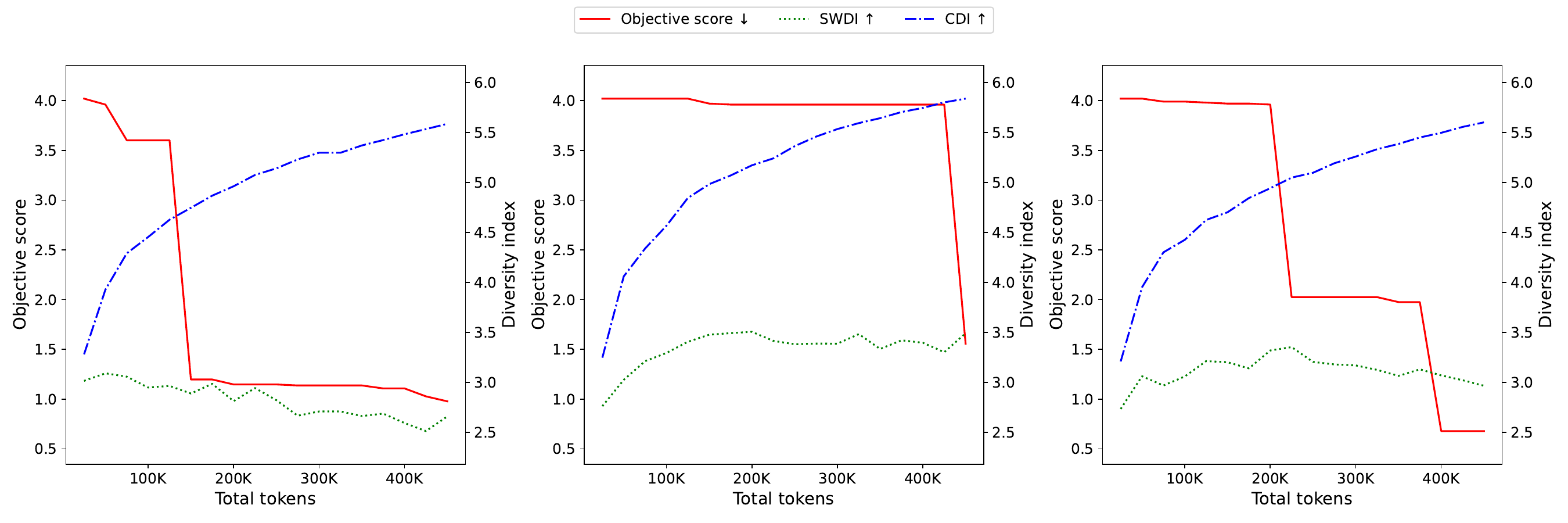} % Reduce the figure size so that it is slightly narrower than the column.
\caption{Diversity indices and objective scores of HSEvo framework on BPO problem through different runs.}
\label{fig:hsevo_di}
\end{figure*}

\section{Experiments}
\subsection{Experimental Setup}

\textbf{Benchmarks:} To assess the diversity and objective scores of HSEvo in comparisons with previous LLM-EPS frameworks, we adopt the same benchmarks as Section \ref{ssec:analysis_cdi_swdi_obj} and conduct experiments on three different AHD problems, BPO \cite{seiden2002online}, TSP \cite{hoffman2013traveling} and OP \cite{ye2024deepaco}.

\begin{itemize}
    \item BPO: packing items into bins of fixed capacity in real-time without prior knowledge of future items. In this benchmark, LLM-EPS frameworks need to design deterministic constructive heuristics to solve.
    \item TSP: find the shortest possible route that visits each city exactly once and returns to the starting point. In this setting, LLM-EPS frameworks need to design heuristics to enhance the perturbation phase for the Guided Local Search solver.
    \item OP: find the most efficient path through a set of locations, maximizing the total score collected within a limited time or distance. This setting requires LLM-EPS frameworks to design herustics used by the ACO solver.
\end{itemize}

\textbf{Experiment settings:} All frameworks were executed under identical environmental settings listed in Table \ref{tab:parameters-settings}. The heuristic search processes across the LLM-EPS frameworks and the evaluations of heuristics were conducted using a single core of an Xeon Processors CPU.

\subsection{Experiment results}

Table \ref{tab:hsevo_results} presents our experiment results on all three AHD problems\footnote{Extending FunSearch to solve the OP problem with an ACO solver caused conflicts we could not resolve with reasonable effort.}. From the table, we observe that while our framework still haven't obtained better CDI than EoH, HSEvo is able to achieve the best objective score on all tasks. On BPO, HSEvo outperforms both FunSearch and ReEvo by a huge margin. This highlights the important of our findings from the analysis in Section \ref{ssec:analysis_llm_eps}, where it is crucial to improve diversity in order to optimize the population better. 

To investigate the impact of our framework on the diversity of the population, we plot the diversity metrics and objective score of HSEvo through different runs on BPO problem in Fig. \ref{fig:hsevo_di}. We can draw a similar observation with findings in Section \ref{ssec:analysis_cdi_swdi_obj}, where with a high SWDI and CDI, the objective score can be optimized significantly. One thing to note here is that in HSEvo first run and ReEvo third run in Fig. \ref{fig:di-and-obj}, both have the SWDI decreasing overtime. However, in HSEvo, the SWDI is at around 3.0 when the objective score improve significantly, then only decrease marginally after that. In ReEvo, the objective score improves when SWDI at around 3.0 and 2.7, and the magnitude of the improvement is not as large as HSEvo, which implies the important of diversity in the optimization of the problem and also the impact of our proposed harmony search.

\subsection{Ablation study}
To gain a better understanding on our novel framework, we conduct ablation studies on our proposed components, the harmony search and flash reflection.
\subsubsection{Harmony search analysis}
As harmony search is a vital component in our HSEvo framework, instead of removing it from HSEvo, we conduct an experiment where we add harmony search into ReEvo framework. Table \ref{tab:ablation_hs} presents our experiment results on OP problem. From the results, we can see that HSEvo outperforms both ReEvo and ReEvo with harmony search variant on both objective score and CDI. Here, notice that harmony search can only marginally improve ReEvo. This can be explained that ReEvo does not have any mechanism to promote diversity, therefore it is not benefitted from the advantage of harmony search process.

\subsubsection{Flash reflection analysis}
We also conduct another experiment where we replace the reflection used in ReEvo with our flash reflection component. As our flash reflection is more cost efficient than original reflection, we reduce the number of tokens used for optimization to 150K. Table \ref{tab:ablation_flash_reflection} presents our experiment results on OP problem. The results show that given a smaller number of timestep, ReEvo with flash reflection mechanism can outperform HSEvo in optimization performance while obtain comparable results on CDI. However, when running with a larger number of tokens, ReEvo with flash reflection cannot improve on both objective score and CDI, while HSEvo improve both metrics to 5.67 and -14.62, respectively. This implies that without diversity-promoting mechanism, flash reflection is not enough to improve the optimization process of LLM-EPS.

\section{Conclusion}
In this paper, we highlight the importance of population diversity in LLM-EPS for AHD problems. We propose two diversity measure metrics, SWDI and CDI, and conduct an analysis on the diversity of previous LLM-EPS approaches. We find that previous approaches either lack focus on the diversity of the population or suffered heavily from the diversity and optimization performance trade-off. We also introduce HSEvo, a novel LLM-EPS framework with diversity-driver harmony search and genetic algorithm for AHD problems. Our experiment results show that our framework can maintain a good balance between diversity and optimization performance trade-off. We also perform additional ablation studies to verify the effectiveness of components of our proposed framework. We hope our work can benefit future research in LLM-EPS community.

\section*{Acknowledgments}
This research was funded by Hanoi University of Science and Technology under project code T2024-PC-038.

\bibliography{aaai25}

\clearpage 

\appendix
\section{Benchmark dataset details}
To demonstrate the effectiveness of HSEvo in particular and LLM-EPS in general, we conduct experiments on three different optimization problems, BPO, TSP, and OP. The data configurations and generation setups used in this paper are regarded as challenging and have been used in recent related studies \cite{romera2024funsearch}, \cite{liu2024eoh}, \cite{ye2024reevo}.

\subsection{Bin Packing Online (BPO)}
\textbf{Definition.} The objective of this problem is to assign a collection of items of varying sizes into the minimum number of containers with a fixed capacity of $C$. Our focus is on the online scenario, where items are packed as they arrive, rather than the offline scenario where all items are known in advance.

\textbf{Dataset generation.} We randomly generate five Weibull instances of size 5k with a capacity of $C = 100$ \cite{romera2024funsearch}. The objective score is set as the average $\frac{lb}{n}$ of five instances, where $lb$ represents the lower bound of the optimal number of bins computed \cite{martello1990lower} and $n$ is the number of bins used to pack all the items by the evaluated heuristic.

\textbf{Solver.} LLM-EPS is used to design heuristic functions that are able to solve this problem directly without any external solver.

\subsection{Traveling Salesman Problem (TSP)}
\textbf{Definition.} The Traveling Salesman Problem (TSP) is a classic optimization challenge that seeks the shortest possible route for a salesman to visit each city in a list exactly once and return to the origin city.

\textbf{Dataset generation.} We generate a set of 64 TSP instances with 100 nodes (TSP100). The node locations in these instances are randomly sampled from $[0, 1]^2$ \cite{voudouris1999guided}. This means that the nodes positioned within a square area bounded by (0, 0) and (1, 1). The average gap from the optimal solution, generated by Concorde \cite{tuHu2022analyzing}, is used as the objective score.

\textbf{Solver.} We use Guided Local Search (GLS) as solver for this benchmark. GLS explores the solution space using local search operations guided by heuristics. The idea behind using this solver is to explore the potential of search penalty heuristics for LLM-EPS. In our experiment, we modified the traditional GLS algorithm by including perturbation phases \cite{arnold2019knowledge}, where edges with higher heuristic values are given priority for penalization. Settings parameters are listed in Table \ref{tab:app-problem-size}, and the number of perturbation moves is 1.

\subsection{Orienteering Problem (OP)}
\textbf{Definition.} In the Orienteering Problem (OP), the objective is to maximize the total score obtained by visiting nodes under a maximum tour length constraint.

\textbf{Dataset generation.} We follow the process of DeepACO \cite{ye2024deepaco} during the generation of this synthetic dataset. In each problem instance, we uniformly sample 50 nodes (OP50), including the depot node, from the unit interval $[0,1]^2$. This means that the nodes positioned within a square area bounded by (0, 0) and (1, 1). We also adopt a challenging prize distribution \cite{kool2018attention}:
$$
p_i = \left(1 + \frac{99 \cdot \frac{d_{0i}}{\max_{j=1}^n d_{0j}}}{100}\right),
$$
where $d_{0i}$ represents the distance between the depot and node $i$, and the maximum tour length constraint is 3.

\textbf{Solver.} We use Ant Colony Optimization (ACO) as solver for this benchmark. ACO is an evolutionary algorithm that integrates solution sampling with pheromone trail updates. It employs stochastic solution sampling, which is biased towards more promising solution spaces based on heuristics. We choose the population size of 20 for ACO to solve OP50.

\section{Diversity analysis}
\subsection{Objective scores and diversity measurement metrics}
\label{appendix:analysis_cdi_swdi_obj}
\textbf{Analysis setting.} The diversity evaluation process was conducted with code embedding model used is \texttt{CodeT5+}\footnote{https://huggingface.co/Salesforce/codet5p-110m-embedding} \cite{wang2023codet5+}, and similarity threshold $\alpha = 0.95$ where $\alpha \in [0;1]$. The maximum budget for each run is 450K tokens.

\begin{figure*}[t]
\centering
\includegraphics[width=0.99\textwidth]{ 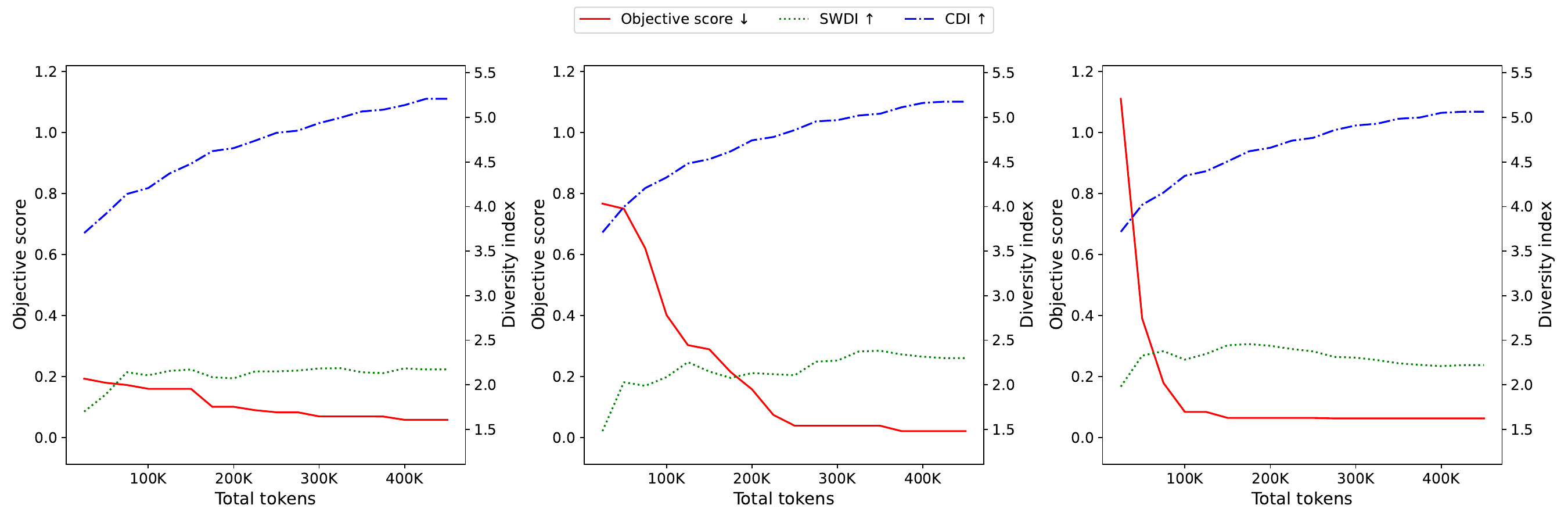} % Reduce the figure size so that it is slightly narrower than the column.
\caption{Diversity indices and objective scores of ReEvo framework on TSP through different runs.}
\label{fig:di-and-obj-2}
\end{figure*}

\begin{figure*}[t]
\centering
\includegraphics[width=0.99\textwidth]{ 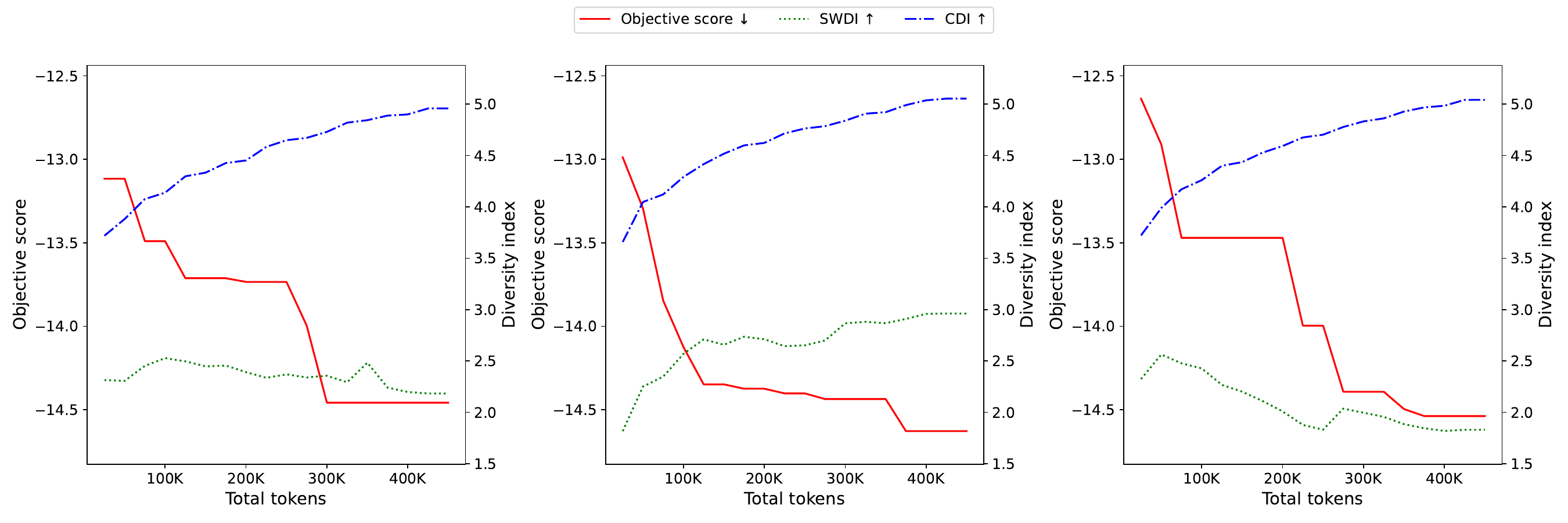} % Reduce the figure size so that it is slightly narrower than the column.
\caption{Diversity indices and objective scores of ReEvo framework on OP through different runs.}
\label{fig:di-and-obj-3}
\end{figure*}

\textbf{Additional results.} The results of the diversity analysis of ReEvo on TSP and OP are shown in the Figure \ref{fig:di-and-obj-2} and Figure \ref{fig:di-and-obj-3}, respectively.

\subsection{Diversity analysis on previous LLM-EPS frameworks}

We conduct experiments with three LLM-EPS frameworks, FunSearch, EoH and ReEvo, on all benchmark datasets, BPO, TSP, and OP. The parameters and LLM information for the corresponding frameworks are presented in Table \ref{tab:parameters-settings}. We follow the same diversity evaluation process as Section \ref{appendix:analysis_cdi_swdi_obj}.

\begin{table}
    \centering
    \begin{tabular}{|c|c|c|c|} \hline 
         Problem& Problem size& \# of iterations& Other\\ \hline 
         BPO& 5000& -& $C$=100\\ \hline
         TSP& 100& 1000& -\\ \hline
         OP& 50& 50& -\\ \hline
    \end{tabular}
    \caption{Problems parameters used for heuristic evaluations}
    \label{tab:app-problem-size}
\end{table}

% \section{HSEvo experiments}

\section{HSEvo prompt examples}
In this section, we synthesize the prompts used in the HSEvo framework. Our prompt system includes four stages initialization population, flash reflection, crossover, and elitist mutation.

\subsection{Task description prompts}
\newcounter{promptcounter}
\stepcounter{promptcounter}
\begin{tcolorbox}[colback=gray!10,colframe=black,title=Prompt \thepromptcounter: System prompt for generator LLM.]
\{role\_init\} helping to design heuristics that can effectively solve optimization problems. 

Your response outputs Python code and nothing else. Format your code as a Python code string: \texttt{```python ... ```}.
\end{tcolorbox}

\stepcounter{promptcounter}
\begin{tcolorbox}[colback=gray!10,colframe=black,title=Prompt \thepromptcounter: Task description.]
\{role\_init\} Your task is to write a \{function\_name\} function for \{problem\_description\}

\{function\_description\}
\end{tcolorbox}

Here, we assign different roles to the LLM  via the variable \texttt{role\_init} to create different personas \cite{shanahan2023role}, which can give LLM different viewpoints and thus encourage the diversity of the population. The roles are selected in a round-robin order from the list: 
\begin{itemize}
    \item \textit{You are an expert in the domain of optimization heuristics,}
    \item \textit{You are Albert Einstein, relativity theory developer,}
    \item \textit{You are Isaac Newton, the father of physics,}
    \item \textit{You are Marie Curie, pioneer in radioactivity,}
    \item \textit{You are Nikola Tesla, master of electricity,}
    \item \textit{You are Galileo Galilei, champion of heliocentrism,}
    \item \textit{You are Stephen Hawking, black hole theorist,}
    \item \textit{You are Richard Feynman, quantum mechanics genius,}
    \item \textit{You are Rosalind Franklin, DNA structure revealer,}
    \item \textit{You are Ada Lovelace, computer programming pioneer.}
\end{itemize}

Variable \texttt{function\_name}, \texttt{problem\_description} and \texttt{function\_description} correspond to the heuristic function name, the specific problem description, and the function description, respectively, which will be detailed in the section \ref{sec:app-prompt-specific}.

\subsection{Population initialization prompt}
\stepcounter{promptcounter}
\begin{tcolorbox}[colback=gray!10,colframe=black,title=Prompt \thepromptcounter: User prompt for population initialization.]
\{task\_description\}

\{seed\_function\}

Refer to the format of a trivial design above. Be very creative and give `\{function\_name\}\_v2`. Output code only and enclose your code with Python code block: \texttt{```python ... ```}.

% \{expert\_knowledge\}
\end{tcolorbox}

Here, \texttt{task\_description} refer to Prompt 2, \texttt{seed\_ function} is a trivial heuristic function for the corresponding problem. 
% In addition, \texttt{expert\_knowledge} is an optional component designed to leverage external expertise in the domain, in all experimental setups we did not use it.

\subsection{Flash reflection prompts}
\stepcounter{promptcounter}
\begin{tcolorbox}[colback=gray!10,colframe=black,title=Prompt \thepromptcounter: System prompt for reflector LLM.]
You are an expert in the domain of optimization heuristics. Your task is to provide useful advice based on analysis to design better heuristics.
\end{tcolorbox}

\stepcounter{promptcounter}
\begin{tcolorbox}[colback=gray!10,colframe=black,title=Prompt \thepromptcounter: User prompt for flash reflection pharse 1.]
\#\#\# List heuristics

Below is a list of design heuristics ranked from best to worst.

\{list\_ranked\_heuristics\}

\#\#\# Guide

- Keep in mind, list of design heuristics ranked from best to worst. Meaning the first function in the list is the best and the last function in the list is the worst.

- The response in Markdown style and nothing else has the following structure:

`**Analysis:**

**Experience:**'

In there:

+ Meticulously analyze comments, docstrings and source code of several pairs (Better code - Worse code) in List heuristics to fill values for **Analysis:**.

Example: ``Comparing (best) vs (worst), we see ...;  (second best) vs (second worst) ...; Comparing (1st) vs (2nd), we see ...; (3rd) vs (4th) ...; Comparing (worst) vs (second worst), we see ...; Overall:...''

+ Self-reflect to extract useful experience for design better heuristics and fill to **Experience:** ($<$ 60 words).

I'm going to tip \$999K for a better heuristics! Let's think step by step.
\end{tcolorbox}

\texttt{list\_ranked\_heuristics} is a list of heuristic functions obtained by grouping parent pairs from selection, removing duplicates, and ranking based on the objective scores.
% from selecting random parent pairs.

\stepcounter{promptcounter}
\begin{tcolorbox}[colback=gray!10,colframe=black,title=Prompt \thepromptcounter: User prompt for flash reflection pharse 2.]
Your task is to redefine `Current self-reflection' paying attention to avoid all things in `Ineffective self-reflection' in order to come up with ideas to design better heuristics.

\#\#\# Current self-reflection

\{current\_reflection\}

\{good\_reflection\}

\#\#\# Ineffective self-reflection

\{bad\_reflection\}

Response ($<$100 words) should have 4 bullet points: Keywords, Advice, Avoid, Explanation.

I'm going to tip \$999K for a better heuristics! Let's think step by step.
\end{tcolorbox}

\texttt{current\_reflection} is the output of the flash reflection step 1 from the current generation. \texttt{good\_reflection} is the output of the flash reflection step 2 from previous generations where a new heuristic was discovered. Conversely, \texttt{bad\_reflection} is the output of the flash reflection step 2 from previous generations where no new heuristic was discovered.

\subsection{Crossover prompt}
\stepcounter{promptcounter}
\begin{tcolorbox}[colback=gray!10,colframe=black,title=Prompt \thepromptcounter: User prompt for crossover.]
\{task\_description\}

\#\#\# Better code

\{function\_signature\_better\}

\{code\_better\}

\#\#\# Worse code

\{function\_signature\_worse\}

\{code\_worse\}

\#\#\# Analyze \& experience

- \{flash\_refection\}

Your task is to write an improved function `\{func\_name\}\_v2` by COMBINING elements of two above heuristics base Analyze \& experience.

Output the code within a Python code block: ```python ... ```, has comment and docstring ($<$50 words) to description key idea of heuristics design.

I'm going to tip \$999K for a better heuristics! Let's think step by step.
\end{tcolorbox}

\texttt{function\_signature\_better}, \texttt{code\_better} and \texttt{function\_signature\_worse}, \texttt{code\_worse} are the function signatures and code of the better and worse parent individuals, respectively. \texttt{flash\_refection} is the output of the flash reflection step 2 from the current generation.

\subsection{Elitist mutation prompt}
\stepcounter{promptcounter}
\begin{tcolorbox}[colback=gray!10,colframe=black,title=Prompt \thepromptcounter: System prompt for elitist mutation.]
\{task\_description\}

Current heuristics:

\{function\_signature\_elitist\}

\{elitist\_code\}

Now, think outside the box write a mutated function `\{func\_name\}\_v2` better than current version. You can using some hints if need:

\{flash\_reflection\}

Output code only and enclose your code with Python code block: \texttt{```python ... ```}.

I'm going to tip \$999K for a better solution!
\end{tcolorbox}

\texttt{function\_signature\_elitist}, \texttt{elitist\_code} are the function signatures and code of elitist individuals of current generation.

\subsection{Harmony Search prompts}

\stepcounter{promptcounter}
\begin{tcolorbox}[colback=gray!10,colframe=black,title=Prompt \thepromptcounter: System prompt for Harmony Search.]
You are an expert in code review. Your task extract all threshold, weight or hardcode variable of the function make it become default parameters.
\end{tcolorbox}

\stepcounter{promptcounter}
\begin{tcolorbox}[colback=gray!10,colframe=black,title=Prompt \thepromptcounter: User prompt for Harmony Search.]
\{elitist\_code\}

Now extract all threshold, weight or hardcode variable of the function make it become default parameters and give me a 'parameter\_ranges' dictionary representation. Key of dict is name of variable. Value of key is a tuple in Python MUST include 2 float elements, first element is begin value, second element is end value corresponding with parameter.

- Output code only and enclose your code with Python code block: \texttt{```python ... ```}.

- Output 'parameter\_ranges' dictionary only and enclose your code with other Python code block: \texttt{```python ... ```}.
\end{tcolorbox}

\texttt{elitist\_code} is the snippet/string code of the elitist individual in the current population that has not yet undergone harmony search.

\subsection{Problem-specific prompts}\label{sec:app-prompt-specific}

Problem-specific prompts are given below:
\begin{itemize}
    \item Table \ref{tab:app-problem-des} presents \texttt{problem\_description} for all problems.
    \item Table \ref{tab:app-func-des} lists the \texttt{function\_description} for all problem settings.
    \item Figure  \ref{fig:app-fun_sig} shows the \texttt{function\_signature}, which also includes the \texttt{function\_name} of each problem.
    \item Figure \ref{fig:app-seed-heuristic} shows the \texttt{seed\_function} used for each problem.
\end{itemize}

\begin{table}[]
    \centering
    \begin{tabular}{|c|p{6cm}|} \hline 
         Problem& Problem description\\ \hline 
         BPO& Solving online Bin Packing Problem (BPP). BPP requires packing a set of items of various sizes into the smallest number of fixed-sized bins. Online BPP requires packing an item as soon as it is received.\\ \hline 
         TSP& Solving Traveling Salesman Problem (TSP) via guided local search. TSP requires finding the shortest path that visits all given nodes and returns to the starting node.\\ \hline 
         OP& Solving a black-box graph combinatorial optimization problem via stochastic solution sampling following ``heuristics".\\ \hline 
    \end{tabular}
    \caption{Problem descriptions used in prompts}
    \label{tab:app-problem-des}
\end{table}

\begin{table}[]
    \centering
    \begin{tabular}{|c|p{6cm}|} \hline 
         Problem& Function description\\ \hline 
         BPO& The priority function takes as input an item and an array of bins\_remain\_cap (containing the remaining capacity of each bin) and returns a priority score for each bin. The bin with the highest priority score will be selected for the item.\\ \hline 
         TSP& The `update\_edge\_distance' function takes as input a matrix of edge distances, a locally optimized tour, and a matrix indicating the number of times each edge has been used. It returns an updated matrix of edge distances that incorporates the effects of the local optimization and edge usage. The returned matrix has the same shape as the input `edge\_distance' matrix, with the distances adjusted based on the provided tour and usage data.\\ \hline 
         OP& The `heuristics' function takes as input a vector of node attributes (shape: n), a matrix of edge attributes (shape: n by n), and a constraint imposed on the sum of edge attributes. A special node is indexed by 0. `heuristics' returns prior indicators of how promising it is to include each edge in a solution. The return is of the same shape as the input matrix of edge attributes. \\ \hline 
    \end{tabular}
    \caption{Function descriptions used in prompts}
    \label{tab:app-func-des}
\end{table}
% Details benchmark, experiments
% prompt example
% HS example 

\begin{figure*}[tb]
\begin{lstlisting}[style=mypython]
# BPO
def priority_v{version}(item: float, bins_remain_cap: np.ndarray) -> np.ndarray:

# TSP
def update_edge_distance_v{version}(edge_distance: np.ndarray, local_opt_tour: np.ndarray, edge_n_used: np.ndarray) -> np.ndarray:

# OP
def heuristics_v{version}(node_attr: np.ndarray, edge_attr: np.ndarray, node_constraint: float)->np.ndarray:
\end{lstlisting}
\caption{Function signatures used in HSEvo.}
\label{fig:app-fun_sig}
\end{figure*}

\begin{figure*}[tb]
\begin{lstlisting}[style=mypython]
# BPO
def priority_v1(item: float, bins_remain_cap: np.ndarray) -> np.ndarray:
    """Returns priority with which we want to add item to each bin.

    Args:
        item: Size of item to be added to the bin.
        bins_remain_cap: Array of capacities for each bin.

    Return:
        Array of same size as bins_remain_cap with priority score of each bin.
    """
    ratios = item / bins_remain_cap
    log_ratios = np.log(ratios)
    priorities = -log_ratios
    return priorities

# TSP
def update_edge_distance(edge_distance: np.ndarray, local_opt_tour: np.ndarray, edge_n_used: np.ndarray) -> np.ndarray:
    """
    Args:
        edge_distance (np.ndarray): Original edge distance matrix.
        local_opt_tour (np.ndarray): Local optimal solution path.
        edge_n_used (np.ndarray): Matrix representing the number of times each edge is used.
    Return:
        updated_edge_distance: updated score of each edge distance matrix.
    """

    num_nodes = edge_distance.shape[0]
    updated_edge_distance = np.copy(edge_distance)

    for i in range(num_nodes - 1):
        current_node = local_opt_tour[i]
        next_node = local_opt_tour[i + 1]
        updated_edge_distance[current_node, next_node] *= (1 + edge_n_used[current_node, next_node])

    updated_edge_distance[local_opt_tour[-1], local_opt_tour[0]] *= (1 + edge_n_used[local_opt_tour[-1], local_opt_tour[0]])
    return updated_edge_distance

# OP
def heuristics_v1(node_attr: np.ndarray, edge_attr: np.ndarray, edge_constraint: float)->np.ndarray:
    return np.ones_like(edge_attr)
\end{lstlisting}
\caption{Seed heuristics used in HSEvo.}
\label{fig:app-seed-heuristic}
\end{figure*}

\section{Generated heuristics}
The best heuristics generated by HSEvo for all problem settings are presented in Figure \ref{fig:app-best-bpo}, \ref{fig:app-best-tsp}, and \ref{fig:app-best-op}.

\begin{figure*}[tb]
\begin{lstlisting}[style=mypython]
import numpy as np
import random
import math
import scipy
import torch
def priority_v2(item: float, bins_remain_cap: np.ndarray, overflow_threshold: float = 1.0987600915713542, mild_penalty: float = 0.5567025232550017, adaptability_lower: float = 0.7264590977149653, adaptability_higher: float = 1.9441643982922379) -> np.ndarray:
    """Enhanced dynamic scoring function for optimal bin selection in online BPP with a more holistic approach."""
    
    # Avoid division by zero by adjusting remaining capacities
    adjusted_bins_remain_cap = np.maximum(bins_remain_cap, np.finfo(float).eps)
    
    # Calculate effective capacities
    effective_cap = np.clip(bins_remain_cap - item, 0, None)
    valid_bins = effective_cap >= 0
    
    # Calculate occupancy ratios with controlled overflow representation
    occupancy_ratio = item / adjusted_bins_remain_cap
    occupancy_scores = np.where(valid_bins, occupancy_ratio, 0)

    # Enhanced overflow penalty: stronger influence for near-overflows
    overflow_penalty = np.where(occupancy_ratio > overflow_threshold, mild_penalty * (occupancy_ratio - overflow_threshold + 1), 1.0)

    # Logarithmic penalty for remaining capacity to encourage SPACE utilization
    log_penalty = np.where(bins_remain_cap > 0, np.log1p(adjusted_bins_remain_cap / (adjusted_bins_remain_cap - item)), 0)

    # Adaptability based on remaining mean capacity
    remaining_mean = np.mean(bins_remain_cap[bins_remain_cap > 0])
    adaptability_factor = np.where(bins_remain_cap < remaining_mean, adaptability_lower, adaptability_higher)
    
    # Comprehensive scoring integrating all metrics for a robust approach
    scores = np.where(valid_bins, occupancy_scores * overflow_penalty * log_penalty * adaptability_factor, -np.inf)
    
    # Normalize scores for prioritization
    max_score = np.max(scores)
    prioritized_scores = (scores - np.min(scores)) / (max_score - np.min(scores)) if max_score > np.min(scores) else scores
    
    # Invert scores for selecting the highest priority bin
    inverted_priorities = 1 - prioritized_scores

    return inverted_priorities
\end{lstlisting}
\caption{The best heuristic for BPO found by HSEvo.}
\label{fig:app-best-bpo}
\end{figure*}

\begin{figure*}[tb]
\begin{lstlisting}[style=mypython]
import numpy as np

def update_edge_distance_v2(edge_distance: np.ndarray, local_opt_tour: np.ndarray, edge_n_used: np.ndarray,
                             penalty_factor: float = 0.6713404008357979, bonus_factor: float = 1.343302294236627, decay_factor: float = 0.3821795974433295,
                             scaling_factor: float = 1.116349420562543, min_distance: float = 7.965736386169868e-05, 
                             penalty_threshold: float = 29.850922399224466, boost_threshold: float = 70.53785604399908) -> np.ndarray:
    """
    Update edge distances using adaptive penalties and bonuses, considering edge usage dynamics
    while ensuring nuanced performance in line with real-time data patterns.
    """
    num_nodes = edge_distance.shape[0]
    updated_edge_distance = np.copy(edge_distance)

    # Calculate average usage for dynamic adjustments
    avg_usage = np.mean(edge_n_used)

    for i in range(num_nodes):
        current_node = local_opt_tour[i]
        next_node = local_opt_tour[(i + 1) % num_nodes]
        usage_count = edge_n_used[current_node, next_node]

        # Adaptive penalty for overused edges
        if usage_count > penalty_threshold:
            # Penalty increases exponentially with usage
            penalty = penalty_factor * (usage_count - penalty_threshold) ** 2
            updated_edge_distance[current_node, next_node] += penalty
            updated_edge_distance[next_node, current_node] += penalty  # Ensure symmetry

        elif usage_count < boost_threshold:
            # Apply a bonus to underused edges
            boost = bonus_factor * (1 + (boost_threshold - usage_count) * 0.1)
            updated_edge_distance[current_node, next_node] *= boost
            updated_edge_distance[next_node, current_node] *= boost  # Ensure symmetry

        # Dynamic scaling based on average usage
        if usage_count > avg_usage:
            adjustment_factor = scaling_factor / (1 + decay_factor ** (usage_count - avg_usage))
        else:
            adjustment_factor = scaling_factor * (1 + decay_factor ** (avg_usage - usage_count))

        # Update the distance with adjustment and ensure non-negative distances
        updated_edge_distance[current_node, next_node] = max(min_distance,
            updated_edge_distance[current_node, next_node] * adjustment_factor)

    return updated_edge_distance
\end{lstlisting}
\caption{The best heuristic for TSP found by HSEvo.}
\label{fig:app-best-tsp}
\end{figure*}

\begin{figure*}[tb]
\begin{lstlisting}[style=mypython]
import numpy as np

def heuristics_v2(node_attr: np.ndarray, edge_attr: np.ndarray, node_constraint: float) -> np.ndarray:
    """
    Enhanced heuristics incorporating contextual adjustments, multi-dimensional scoring,
    and adaptive responsiveness to edge conditions.
    """
    normalization_epsilon = 1e-8
    influence_threshold = 0.9
    adaptability_factor = 2.0
    non_linearity_base = 2.5
    n = node_attr.shape[0]
    score_matrix = np.zeros_like(edge_attr)

    # Normalize node attributes (maintain zero division safety)
    total_node_attr = np.sum(node_attr) + normalization_epsilon
    normalized_node_attr = node_attr / total_node_attr

    for i in range(n):
        for j in range(n):
            if i == j:
                continue  # Skip self-loops

            # Calculate contextual adjustment for edge attributes
            dynamic_edge = edge_attr[i, j] ** adaptability_factor
            adjusted_edge = dynamic_edge / (node_constraint + normalization_epsilon)

            # Multi-dimensional scaling based on edge and node attributes
            if adjusted_edge > influence_threshold:
                scaling_factor = np.sqrt(adjusted_edge + normalization_epsilon)
            else:
                scaling_factor = influence_threshold / (adjusted_edge + normalization_epsilon)

            # Layered logic for combined node influence using non-linear model
            combined_node_influence = (
                (normalized_node_attr[i] ** non_linearity_base) +
                (normalized_node_attr[j] ** non_linearity_base)
            ) ** 1.5  # Further amplify the influence

            # Score calculation with contextual penalties (non-linearity)
            score = combined_node_influence * scaling_factor
            
            if dynamic_edge > 0:
                penalty_base = np.log1p(dynamic_edge)  # Non-linear penalty
                penalty_factor = 1 / (1 + penalty_base ** 2)  # Stronger sensitivity with edges
                score *= penalty_factor
            
            # Normalization to retain meaningful scale
            score_matrix[i, j] = score / (dynamic_edge + normalization_epsilon)

    return score_matrix
\end{lstlisting}
\caption{The best heuristic for OP found by HSEvo.}
\label{fig:app-best-op}
\end{figure*}

\end{document}